\documentclass[11pt,a4paper]{article}

\usepackage[hyperref]{acl2018}
\usepackage{times}
\usepackage{latexsym}
\usepackage{url}
\usepackage{amsmath}
\usepackage{amssymb}
\usepackage{amsfonts}
\usepackage{graphicx}
\usepackage[skip=5pt]{caption}
\usepackage{subcaption}
\usepackage{array}
\usepackage{tabu}
\usepackage{makecell}
\usepackage{subcaption}
\usepackage{paralist}
\usepackage{cases}
\usepackage{diagbox}
\usepackage{enumitem}
\usepackage{soul}
\usepackage{multirow}
\usepackage{verbatim}
\usepackage{tabulary}
\usepackage{booktabs}
\usepackage[mathscr]{euscript}
\usepackage{mathtools}
\usepackage{amssymb}
\usepackage{algorithm}
\usepackage{algpseudocode}
\algrenewcommand{\algorithmiccomment}[1]{\leavevmode\hfill$\triangleright$ #1}
\usepackage{color}
\usepackage{bm}

\definecolor{orange}{rgb}{1,0.5,0}
\definecolor{mdgreen}{rgb}{0.05,0.6,0.05}
\definecolor{mdblue}{rgb}{0,0,0.7}
\definecolor{dkblue}{rgb}{0,0,0.5}
\definecolor{dkgray}{rgb}{0.3,0.3,0.3}
\definecolor{slate}{rgb}{0.25,0.25,0.4}
\definecolor{gray}{rgb}{0.5,0.5,0.5}
\definecolor{ltgray}{rgb}{0.7,0.7,0.7}
\definecolor{purple}{rgb}{0.7,0,1.0}
\definecolor{lavender}{rgb}{0.65,0.55,1.0}

\newcommand{\ensuretext}[1]{#1}
\newcommand{\marker}[2]{\ensuremath{^{\textsc{#1}}_{\textsc{#2}}}}

\newcommand{\arkcomment}[3]{\ensuretext{\textcolor{#3}{[#1 #2]}}}
\renewcommand{\arkcomment}[3]{}  

\newcommand{\nascomment}[1]{\arkcomment{\marker{NA}{S}}{#1}{blue}}

\newcommand{\com}[1]{}

\newcommand{\term}[1]{\textbf{#1}} 

\newcommand{\ledge}[3]{#1 \overset{#3}{\rightarrow} #2}
\newcommand{\edge}[2]{#1 {\rightarrow} #2}

\newcommand{\bilstm}{\mathbf{h}}
\newcommand{\fw}[1]{\overrightarrow{\bilstm}_{#1}}
\newcommand{\bw}[1]{\overleftarrow{\bilstm}_{#1}}

\newcommand{\interalia}[1]{\citep[\emph{inter alia}]{#1}}
\newcommand{\codeurl}{\url{https://github.com/Noahs-ARK/SPIGOT}}

\newcommand{\argmax}[1]{\underset{#1}{\operatorname{argmax}}\;}

\newcommand{\argmaxinline}[1]{\operatorname{argmax}_{#1}}
\newcommand{\argmininline}[1]{\operatorname{argmin}_{#1}}
\newcommand{\argmaxname}{\operatorname{argmax}}

\newcommand{\softmax}{\operatorname{softmax}}
\newcommand{\rulesep}{\unskip\ \vrule\ }

\usepackage{xspace}

\newcommand{\spigot}{\textsc{spigot}\xspace}
\newcommand{\pfgd}{\spigot}
\newcommand{\longSpigot}{structured projection of intermediate gradients\xspace}
\newcommand{\longPfgd}{\longSpigot}
\newcommand{\ste}{\textsc{ste}\xspace}
\newcommand{\sa}{\textsc{sa}\xspace}
\newcommand{\pipeline}{\textsc{pipeline}\xspace}

\newcommand{\norm}[1]{\left\lVert#1\right\rVert}

\newcolumntype{L}[1]{>{\raggedright\let\newline\\\arraybackslash\hspace{0pt}}m{#1}}
\newcolumntype{C}[1]{>{\centering\let\newline\\\arraybackslash\hspace{0pt}}m{#1}}
\newcolumntype{R}[1]{>{\raggedleft\let\newline\\\arraybackslash\hspace{0pt}}m{#1}}

\setul{1pt}{.4pt}

\DeclareCaptionFont{tiny}{\tiny}
\aclfinalcopy 
\def\aclpaperid{1248} 

\title{Backpropagating through Structured Argmax using a \spigot}

\author{
	Hao Peng$^\diamondsuit$ \quad
	Sam Thomson$^\clubsuit$ \quad
	Noah A. Smith$^\diamondsuit$ \\
	$^\diamondsuit$ Paul G. Allen School of Computer Science \& Engineering,
	University of Washington\\
	$^\clubsuit$ School of Computer Science,
	Carnegie Mellon University\\ 
	{\tt \{hapeng,nasmith\}@cs.washington.edu,
	sthomson@cs.cmu.edu}
}
\date{}

\begin{document}
\maketitle

\begin{abstract}
We introduce the \term{\longPfgd} optimization technique~(\pfgd), a
new method for backpropagating through neural networks that include
hard-decision structured predictions (e.g., parsing) in intermediate
layers.~\pfgd requires no marginal inference, unlike structured
attention networks~\citep{kim2017structured} and some reinforcement learning-inspired
solutions~\citep{yogatama2016learning}.
Like so-called straight-through estimators \citep{hinton2012neural}, 
\pfgd defines gradient-like quantities associated with
intermediate nondifferentiable operations, allowing backpropagation
before and after them; \pfgd's proxy aims to ensure that, after
a parameter update, the intermediate structure will remain well-formed.

We experiment on two structured NLP pipelines: syntactic-then-semantic
dependency parsing, and semantic parsing
followed by sentiment classification.
We show that training with \pfgd leads to a larger improvement on the downstream
task than a modularly-trained pipeline, the straight-through estimator,
and structured attention,
reaching a new state of the art on semantic dependency parsing.
\end{abstract}

\section{Introduction}
\label{sec:intro}

Learning methods for natural language processing are increasingly
dominated by end-to-end differentiable functions that can be trained
using gradient-based optimization. Yet traditional NLP often assumed
modular stages of processing that formed a pipeline; e.g.,
text was tokenized, then tagged with parts of speech, then parsed into
a phrase-structure or dependency tree, then semantically analyzed.
Pipelines, which make ``hard'' (i.e., discrete) decisions at each
stage, appear to be incompatible with neural learning, leading many
researchers to abandon earlier-stage processing.

Inspired by findings that continue to see benefit from various kinds
of linguistic or domain-specific preprocessing~\citep{he2017deep,oepen2017epe,ji2017neural}, 
we argue that pipelines can be
treated as layers in neural architectures for NLP tasks.
Several solutions are readily available:
\begin{compactitem}
\item Reinforcement learning (most notably the \textsc{reinforce}
  algorithm;~\citealp{williams1992refinforce}),
  and \textbf{structured attention} (\sa; \citealp{kim2017structured}).
  These methods replace $\argmaxname$
  with a sampling or marginalization operation.
  We note two potential downsides of these approaches:
  (i) not all $\argmaxname$-able operations have corresponding sampling
  or marginalization methods that are efficient, and (ii) inspection
  of intermediate outputs, which could benefit error analysis and
  system improvement, is more straightforward for hard decisions
  than for posteriors.
\item The \textbf{straight-through estimator} (\ste; \citealp{hinton2012neural})
  treats discrete decisions as if they were differentiable and
  simply passes through gradients.
  While fast and surprisingly effective, it ignores
  \emph{constraints} on the $\argmaxname$ problem, such as the requirement
  that every word has exactly one syntactic parent. 
  We will find, experimentally, that the quality of
  intermediate representations degrades substantially under \ste.
\end{compactitem}
This paper introduces a new method, the \term{\longSpigot} optimization technique~(\spigot;
\S\ref{sec:method}), which defines a proxy for the gradient of a loss
function with respect to the input to $\argmaxname$.  Unlike \ste's gradient proxy,
\spigot aims to respect the constraints in the $\argmaxname$ problem.
\spigot can be applied with any intermediate layer that is expressible
as a constrained maximization problem, and whose feasible set can be projected
onto.
We show empirically that \spigot works even when the maximization and the projection are
done approximately.

We offer two concrete architectures that employ structured $\argmaxname$ 
as an intermediate layer: semantic parsing with
syntactic parsing in the middle,
and sentiment analysis with semantic parsing in the middle~(\S\ref{sec:projection}). 
These architectures are trained using a joint
objective, with one part using data for the intermediate task, and the
other using data for the end task.
The datasets are not assumed to overlap at all, but the parameters for the
intermediate task are affected by both parts of the training data.

Our experiments (\S\ref{sec:experiment}) show that our architecture improves over a
state-of-the-art semantic dependency parser, and that \spigot offers
stronger performance than a pipeline, \sa, and \ste.  On
sentiment classification, we show that semantic parsing offers
improvement over a BiLSTM, more so with \spigot than with
alternatives.
Our analysis considers how the behavior of the intermediate parser is
affected by the end task (\S\ref{sec:analysis}).
Our code is open-source and available at \codeurl.

\section{Method}
\label{sec:method}

Our aim is to allow a (structured) $\argmaxname$ layer in a neural
network to be treated almost like any other differentiable function.  This would
allow us to place, for example, a syntactic parser in the middle of a
neural network, so that the forward calculation simply calls the
parser and passes the parse tree to the next layer, which might derive
syntactic features for the next stage of processing.  

The challenge is in the \emph{backward} computation, which is key to
learning with standard gradient-based methods.
When its output is discrete as we
assume here, $\argmaxname$ is a piecewise constant function.
At every point, its gradient is either zero or undefined.
So instead of using the true gradient, we will introduce a \emph{proxy} for the
gradient of the loss function with respect to the inputs to
$\argmaxname$, allowing backpropagation to proceed through the
$\argmaxname$ layer.  Our proxy is designed as an improvement to
earlier methods (discussed below) that completely ignore constraints
on the $\argmaxname$ operation.  It accomplishes this through a
projection of the gradients.

We first lay out notation, 
and then briefly review max-decoding and its relaxation (\S\ref{subsec:max-decoding}).
We define \pfgd in \S\ref{subsec:pfgd},
and show how to use it to backpropagate through NLP pipelines in \S\ref{subsec:backprop}.

\paragraph{Notation.}
Our discussion centers around two tasks:
a structured \emph{intermediate} task followed by an \emph{end} task,
where the latter considers the outputs of the former (e.g., syntactic-then-semantic parsing).
Inputs are denoted as $\mathbf{x}$, and end task outputs as $\mathbf{y}$.
We use $\mathbf{z}$ to denote intermediate structures derived from $\mathbf{x}$.
We will often refer to the intermediate task as ``decoding'', in the structured prediction sense.
It seeks an output $\hat{\mathbf{z}}=\argmaxinline{\mathbf{z}\in\mathcal{Z}}S$ 
from the feasible set $\mathcal{Z}$, maximizing a (learned, parameterized) scoring function $S$
for the structured intermediate task.
$L$ denotes the loss of the end task, which may or may not also involve structured predictions.
We use
$\Delta^{k-1}=\{\mathbf{p}\in\mathbb{R}^{k}\mid\mathbf{1}^\top\mathbf{p}=1, \mathbf{p}\geq\mathbf{0}\}$
to denote the ($k-1$)-dimensional simplex.
We denote the domain of binary variables as $\mathbb{B} = \{0, 1\}$,
and the unit interval as $\mathbb{U} = [0,1]$.
By projection of a vector $\mathbf{v}$ onto a set $\mathcal{A}$, 
we mean the closest point in $\mathcal{A}$ to $\mathbf{v}$,
measured by Euclidean distance:
$\operatorname{proj}_{\mathcal{A}}(\mathbf{v}) =
\argmininline{\mathbf{v}^\prime \in \mathcal{A}}{ \norm{\mathbf{v}^\prime - \mathbf{v}}_2 }$.

\subsection{Relaxed Decoding}
\label{subsec:max-decoding}
Decoding problems 
are
typically decomposed into a collection of ``parts'', such as arcs in a
dependency tree or graph.  In such a setup, each element of
$\mathbf{z}$, $z_i$, corresponds to one possible part, and $z_i$ takes
a boolean value to indicate whether the part is included in the output
structure.  The scoring function $S$ is assumed to decompose into a
vector $\mathbf{s}(\mathbf{x})$ of part-local, input-specific scores:
\begin{align}
\label{eq:score}
\hat{\mathbf{z}} 
= \argmax{\mathbf{z}\in\mathcal{Z}}{S(\mathbf{x},\mathbf{z})}
= \argmax{\mathbf{z}\in\mathcal{Z}}{\mathbf{z}^\top\mathbf{s}(\mathbf{x})}
\end{align}
In the following, we drop $\mathbf{s}$'s dependence on $\mathbf{x}$
for clarity.

In many NLP problems, the output space $\mathcal{Z}$ can be specified 
by linear constraints~\citep{roth2004linear}:
\begin{align}
\label{eq:constraints}
\mathbf{A}
\begin{bmatrix}
\mathbf{z}\\
\bm{\psi}
\end{bmatrix}
\leq\mathbf{b},
\end{align}
where $\bm{\psi}$ are auxiliary variables (also scoped by $\argmaxname$),
together with integer constraints (typically, each $z_i \in \mathbb{B}$).

The problem in Equation~\ref{eq:score} can be NP-complete in general,
so the $\{0,1\}$ constraints are often relaxed
to $[0,1]$ to make decoding tractable~\citep{martins2009polyhedral}.
Then the discrete combinatorial problem over $\mathcal{Z}$ is transformed
into the optimization of a linear objective over a convex polytope
$\mathcal{P}\hspace{-.1cm}=\hspace{-.1cm}\{\mathbf{p}\in\mathbb{R}^d
\hspace{-.1cm}\mid\hspace{-.1cm}\mathbf{A}\mathbf{p}\hspace{-.1cm}\leq\hspace{-.1cm}\mathbf{b}\}$,
which is solvable in polynomial time~\citep{bertsimas1997lp}. 
This is not necessary in some cases, where the $\argmaxname$ can be solved exactly
with dynamic programming.




\begin{figure}
	\centering
	\begin{subfigure}[b]{.48\columnwidth}
		\centering
		\includegraphics[clip,trim=10cm 9.5cm 11.5cm 9.5cm, width=\columnwidth]{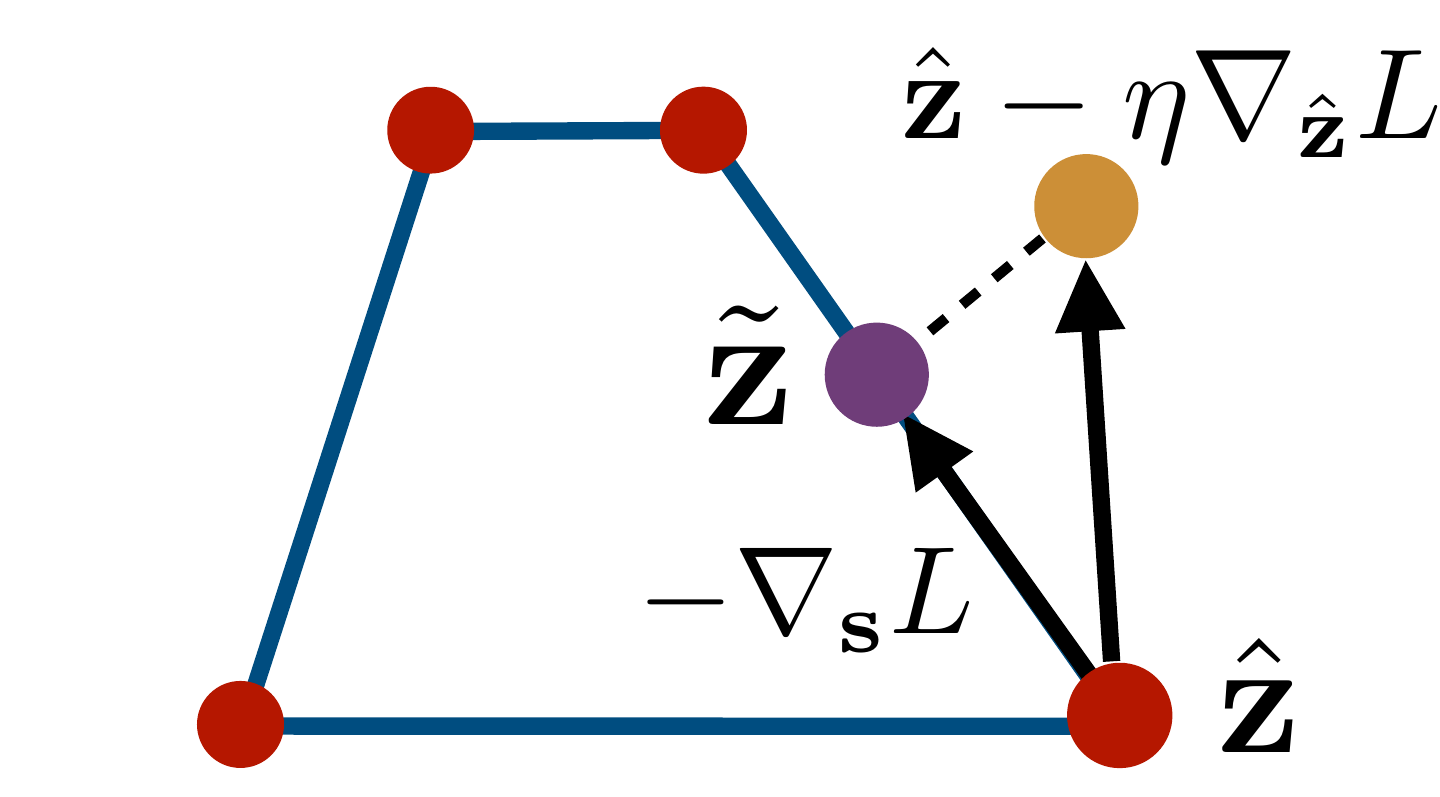}
	\end{subfigure}
	\rulesep
	\begin{subfigure}[b]{.48\columnwidth}
		\centering
		\includegraphics[clip,trim=10cm 9.5cm 11.5cm 9.5cm, width=\columnwidth]{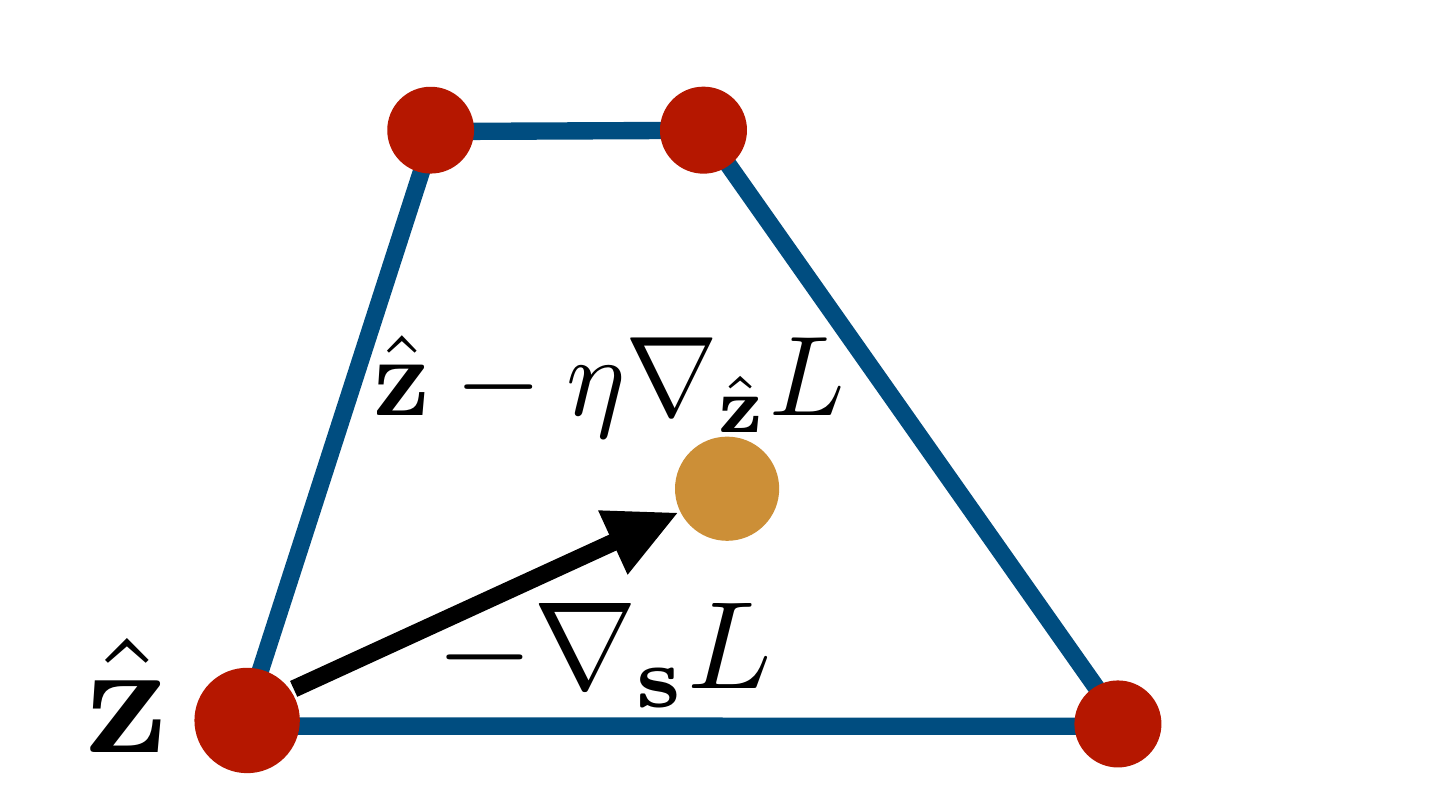}
	\end{subfigure}
	\caption{The original feasible set $\mathcal{Z}$ (red vertices),
		is relaxed into a convex polytope $\mathcal{P}$ (the area encompassed by blue edges).
		Left: making a gradient update to $\hat{\mathbf{z}}$
		makes it step outside the polytope, and it is projected back to $\mathcal{P}$, 
		resulting in the projected point $\tilde{\mathbf{z}}$. 
		$\nabla_{\mathbf{s}}L$ is then along the edge.
		Right: updating $\hat{\mathbf{z}}$ keeps it within $\mathcal{P}$,
		and thus $\nabla_{\mathbf{s}}L=\eta\nabla_{\hat{\mathbf{z}}}L$.
		\label{fig:pfgd}}
\end{figure}

\subsection{From \ste\ to \spigot}
\label{subsec:pfgd}
We now view structured $\argmaxname$ as an activation function
that takes a vector of input-specific part-scores $\mathbf{s}$ and
outputs a solution $\hat{\mathbf{z}}$.  For backpropagation, to
calculate gradients for
parameters of $\mathbf{s}$, the chain rule defines:
\begin{equation}
\label{eq:ste}
\nabla_{\mathbf{s}}L = \mathbf{J}\ \nabla_{\hat{\mathbf{z}}} L,
\end{equation}
where the Jacobian matrix 
$\mathbf{J}\hspace{-.05cm}=\hspace{-.05cm}\frac{\partial\hat{\mathbf{z}}}{\partial\mathbf{s}}$
contains the derivative of each
element of $\hat{\mathbf{z}}$ with respect to each element of $\mathbf{s}$.
Unfortunately, $\argmaxname$ is a piecewise constant function, so its Jacobian 
is either zero (almost everywhere) or undefined (in the case of ties).

One solution, taken in \emph{structured attention}, 
is to replace the $\argmaxname$
with marginal inference and a $\softmax$ function, so that $\hat{\mathbf{z}}$ encodes 
probability distributions over parts \citep{kim2017structured,liu2017learning}.  As discussed in
\S\ref{sec:intro}, there are two reasons to avoid this
modification.
Softmax can only be used when marginal inference is feasible, by sum-product algorithms for example
\citep{eisner2016inside,Friesen2016TheST}; in general marginal inference can be
\#P-complete.  Further, a soft intermediate layer will be less
amenable to inspection by anyone wishing to understand and improve the model.

In another line of work,
$\argmaxname$ is augmented with a strongly-convex penalty on the
solutions~\citep{martins2016sparsemax,amos2017optnet,niculae2017regularized,niculae2018sparsemap,mensch2018differentiable}.
However, their approaches require solving a relaxation even when 
exact decoding is tractable.
Also, the penalty will bias the solutions found by the decoder,
which may be an undesirable conflation of computational and modeling concerns.

A simpler solution is the \ste method~\citep{hinton2012neural}, which
replaces the Jacobian matrix in Equation~\ref{eq:ste} by the identity matrix.  This method has
been demonstrated to work  well when used to ``backpropagate'' through 
hard threshold functions~\citep{bengio_estimating_2013,friesen2017deep}
and categorical random
variables~\citep{jang2016categorical,choi2017unsupervised}.

Consider for a moment what we would do if $\hat{\mathbf{z}}$ were
a vector of parameters, rather than intermediate predictions.
In this case, we are
seeking points in $\mathcal{Z}$ that minimize $L$; denote that set
of minimizers by $\mathcal{Z}^\ast$.
Given $\nabla_{\hat{\mathbf{z}}}L$ and step size $\eta$, we would update $\hat{\mathbf{z}}$
to be $\hat{\mathbf{z}} - \eta \nabla_{\hat{\mathbf{z}}}L$.  This
update, however, might not return a value in the feasible set
$\mathcal{Z}$, or even (if we are using a linear relaxation) the
relaxed set $\mathcal{P}$.  

\spigot therefore introduces a \emph{projection} step that aims to
keep the ``updated'' $\hat{\mathbf{z}}$ in the feasible set.  Of
course, we do not directly update $\hat{\mathbf{z}}$; we continue
backpropagation through $\mathbf{s}$ and onward to the parameters.
But the projection step nonetheless alters the parameter updates in
the way that our proxy for ``$\nabla_{\mathbf{s}}L$''  is defined.

The procedure is defined as follows: 
\begin{subequations}
	\begin{align}
	\mathbf{\hat{p}} &= \hat{\mathbf{z}} -
                           \eta\nabla_{\hat{\mathbf{z}}}L, \label{eq:pr}\\
	\tilde{\mathbf{z}} &=
	\operatorname{proj}_{\mathcal{P}}(\mathbf{\hat{p}}) \label{eq:proj},\\
	\nabla_{\mathbf{s}}L&\triangleq \hat{\mathbf{z}} - \tilde{\mathbf{z}}\label{eq:proj_grad}.
	\end{align}%
	\label{eq:pfgd}%
\end{subequations}%
First, the method makes an ``update'' to $\hat{\mathbf{z}}$ as if it
contained parameters (Equation~\ref{eq:pr}), letting $\mathbf{\hat{p}}$
denote the new value. Next, $\mathbf{\hat{p}}$ is projected back onto
the (relaxed) feasible set (Equation~\ref{eq:proj}), yielding a feasible
new value $\tilde{\mathbf{z}}$.  
Finally, the gradients with respect to $\mathbf{s}$ are computed by
 Equation~\ref{eq:proj_grad}.


Due to the convexity of $\mathcal{P}$,
the projected point $\tilde{\mathbf{z}}$ will always be unique,
and is guaranteed to be no farther than $\mathbf{\hat{p}}$
from any point in $\mathcal{Z}^{\ast}$~\citep{luenberger2015linear}.%
\footnote{Note that this property follows from $\mathcal{P}$'s convexity,
and we do not assume the convexity of $L$.} 
Compared to \ste, \pfgd~involves a projection and limits
$\nabla_\mathbf{s} L$
to a smaller space to satisfy constraints.
See Figure~\ref{fig:pfgd} for an illustration.

When efficient exact solutions (such as dynamic
programming) are available, they can be used.
Yet, we note that \spigot does not assume the $\argmaxname$ operation is
solved exactly.

\subsection{Backpropagation through Pipelines}
\label{subsec:backprop}

Using \spigot, we now devise an algorithm to ``backpropagate'' through
NLP pipelines.
In these pipelines, an intermediate task's output is
fed into an end task for use as features.
The parameters 
of the complete model are divided into two parts:
denote the parameters of the intermediate task model by $\bm{\phi}$
(used to calculate $\mathbf{s}$),
and those in the end task model as $\bm{\theta}$.\footnote{Nothing
  prohibits tying across pre-$\argmaxname$ parameters and
  post-$\argmaxname$ parameters; this separation is notationally
  convenient but not at all necessary.}
As introduced earlier, the end-task loss function to be minimized is $L$, which
depends on both $\bm{\phi}$ and $\bm{\theta}$.

Algorithm~\ref{algo:pfgd} describes the forward and backward computations.
It takes an end task training pair $\langle \mathbf{x},\mathbf{y}\rangle$,
along with the intermediate task's feasible set $\mathcal{Z}$, which is determined by $\mathbf{x}$.
It first runs the intermediate model and decodes to get intermediate structure $\hat{\mathbf{z}}$,
just as in a standard pipeline.
Then forward propagation is continued into the end-task model to
compute loss 
$L$, using $\hat{\mathbf{z}}$ to define input features.
Backpropagation in the end-task model computes $\nabla_{\bm{\theta}}L$
and $\nabla_{\hat{\mathbf{z}}}L$,
and $\nabla_{\mathbf{s}}L$ is then constructed using Equations~\ref{eq:pfgd}.
Backpropagation then continues into the intermediate model, computing $\nabla_{\bm{\phi}}L$.

Due to its flexibility, \pfgd~is applicable to many training scenarios.
When there is no $\langle \mathbf{x}, \mathbf{z}\rangle$ training data for the intermediate task, \pfgd can be used to induce latent structures
for the end-task \interalia{yogatama2016learning,kim2017structured,choi2017unsupervised}.
When intermediate-task training data \emph{is} available,
one can use \pfgd~to adopt joint learning by minimizing an
interpolation of $L$ (on end-task data $\langle \mathbf{x},\mathbf{y}\rangle$) and an intermediate-task loss
function $\widetilde{L}$ (on intermediate
task data $\langle \mathbf{x}, \mathbf{z}\rangle$). 
This is the setting in our experiments; note that we do not assume any overlap in the training examples for the two tasks.


\begin{algorithm}
	\small
	\centering
	\caption{Forward and backward computation with \pfgd.}
	\label{algo:pfgd}
	\begin{algorithmic}[1]
		\Procedure{\pfgd}{$\mathbf{x},\mathbf{y}, \mathcal{Z}$}
		\State Construct $\mathbf{A}$, $\mathbf{b}$ 
		such that $\mathcal{Z} = \{\mathbf{p}\in\mathbb{Z}^d\mid \mathbf{A}\mathbf{p}\leq\mathbf{b}\}$
		\State $\mathcal{P}\leftarrow\{\mathbf{p}\in\mathbb{R}^d\mid \mathbf{A}\mathbf{p}\leq\mathbf{b}\}$
		\Comment Relaxation
		\State Forwardprop and compute $\mathbf{s}_{\bm{\phi}}(\mathbf{x})$
		\State $\hat{\mathbf{z}}\leftarrow
		\argmaxinline{\mathbf{z}\in\mathcal{Z}}{\mathbf{z}^\top\hspace{-.05cm}\mathbf{s}_{\bm{\phi}}
                (\mathbf{x})}$ \Comment Intermediate decoding
		\State Forwardprop and compute
                $L$ given $\mathbf{x}$, $\mathbf{y}$, and $\hat{\mathbf{z}}$
		\State Backprop and compute
                $\nabla_{\bm{\theta}}L$ and $\nabla_{\hat{\mathbf{z}}}L$
		\State $\tilde{\mathbf{z}}\leftarrow \operatorname{proj}_{\mathcal{P}}(
		  \hat{\mathbf{z}} - \eta\nabla_{\hat{\mathbf{z}}}L
		)$ 
		\Comment Projection
		\State $\nabla_{\mathbf{s}}L \leftarrow\hat{\mathbf{z}}-\tilde{\mathbf{z}}$
		\State Backprop and compute $\nabla_{\bm{\phi}}L$
		\EndProcedure
	\end{algorithmic}
\end{algorithm}

\section{Solving the Projections}
\label{sec:projection}

In this section we discuss how to compute approximate projections
for the two intermediate tasks considered in this work,
arc-factored unlabeled dependency parsing
and first-order semantic dependency parsing.

In early experiments we observe that for both tasks,
projecting with respect to \emph{all} constraints of their original formulations
using a generic quadratic program solver was prohibitively slow.
Therefore, we construct relaxed polytopes 
by considering only a subset of the constraints.%
\footnote{A parallel work introduces
an active-set algorithm to solve the same class of quadratic programs~\citep{niculae2018sparsemap}.
It might be an efficient approach to solve the projections in Equation~\ref{eq:proj}, 
which we leave to future work.}
The projection then decomposes into a series of singly constrained quadratic programs (QP),
each of which can be efficiently solved in linear time.

The two approximate projections discussed here are used in backpropagation only.
In the forward pass, we solve the decoding problem using the models' original decoding algorithms.

\paragraph{Arc-factored unlabeled dependency parsing.}
For unlabeled dependency trees, we impose $[0,1]$ constraints and
single-headedness constraints.%
\footnote{
	It requires $O(n^2)$ auxiliary variables and $O(n^3)$ additional constraints
	to ensure well-formed tree structures~\citep{martins2013turboparser}.
}

Formally, given a length-$n$ input sentence, excluding self-loops, 
an arc-factored parser considers $d=n(n-1)$ candidate arcs.
Let $\edge{i}{j}$ denote an arc from the $i$th token to the $j$th,
and $\sigma(\edge{i}{j})$ denote its index. 
We construct the relaxed feasible set by:
\begin{equation}
\label{eq:dep_proj}
\begin{split}
\mathcal{P}_{\text{DEP}} = \left\{\mathbf{p}\in\mathbb{U}^d \left|
\sum_{i\neq j}p_{\sigma(\edge{i}{j})}=1,\forall j\right.\right\},
\end{split}
\end{equation}
i.e., we consider each token $j$ individually,
and force single-headedness by constraining the number of arcs incoming to $j$ to sum to 1.
Algorithm~\ref{algo:dep_proj} summarizes the procedure to project onto $\mathcal{P}_{\text{DEP}}$.
Line~\ref{algo:line:dep_proj_simplex} 
forms a singly constrained QP, and can be solved in $O(n)$ time~\citep{brucker1984qp}.



\begin{figure}
	\centering
	\includegraphics[clip,trim=5cm 10.cm 5.5cm 11.cm, width=\columnwidth]{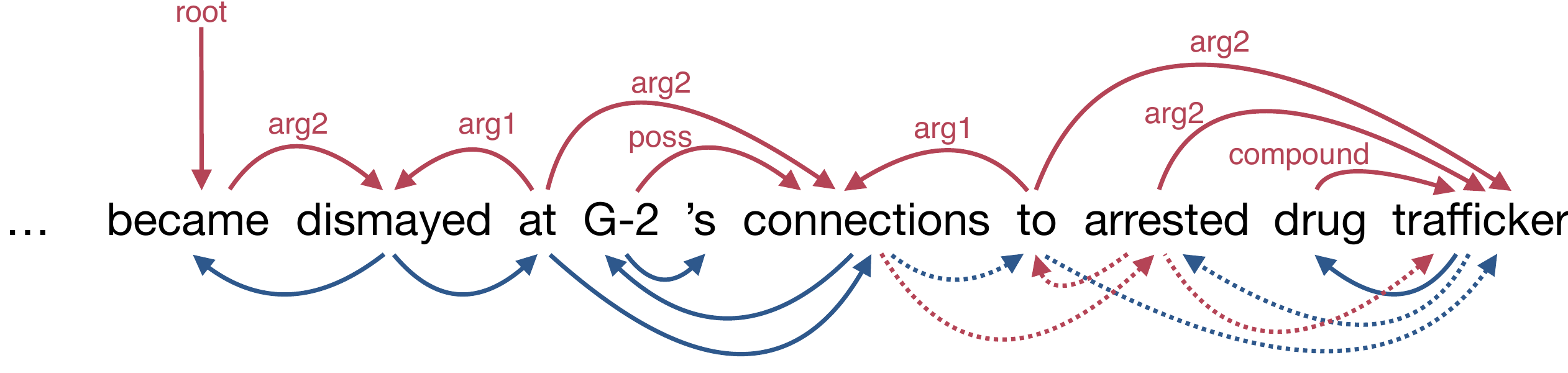}
	\caption{A development instance annotated with 
		both gold DM semantic dependency graph (red arcs on the top),
		and gold syntactic dependency tree (blue arcs at the bottom). 
		A pretrained syntactic parser predicts the same tree as the gold;
		the semantic parser backpropagates 
		into the intermediate syntactic parser,
		and changes the dashed blue arcs into dashed red arcs (\S\ref{sec:analysis}).}
	\label{fig:form}
\end{figure}

\begin{algorithm}
	\small
	\centering
	\caption{Projection onto the relaxed polytope $\mathcal{P}_{\text{DEP}}$
		for dependency tree structures.
		Let bold $\bm{\sigma}(\edge{\cdot}{j})$ denote the index set of arcs incoming to $j$.
		For a vector $\mathbf{v}$, we use $\mathbf{v}_{\bm{\sigma}(\edge{\cdot}{j})}$
		to denote vector $[v_k]_{k\in \bm{\sigma}(\edge{\cdot}{j})}$.
	}
	\label{algo:dep_proj}
	\begin{algorithmic}[1]
		\Procedure{DepProj}{$\hat{\mathbf{p}}$}
		\For{$j=1,2,\dots,n$}
		\State$\tilde{\mathbf{z}}_{\bm{\sigma}(\edge{\cdot}{j})}\leftarrow
		\text{proj}_{\Delta^{n-2}}\left(
		\hat{\mathbf{p}}_{\bm{\sigma}(\edge{\cdot}{j})}
		\right)$  \label{algo:line:dep_proj_simplex}
		\EndFor
		\State\Return $\tilde{\mathbf{z}}$
		\EndProcedure
	\end{algorithmic}
\end{algorithm}
\paragraph{First-order semantic dependency parsing.}
Semantic dependency parsing uses labeled bilexical dependencies
to represent sentence-level semantics~\citep{oepen2014sdp,oepen2015sdp,Oep:Kuh:Miy:16}.
Each dependency is represented by a labeled directed arc from a head token to a modifier token,
where the arc label encodes broadly applicable semantic relations.
Figure~\ref{fig:form} diagrams a semantic graph from the DELPH-IN MRS-derived dependencies (DM),
together with a syntactic tree.

We use a state-of-the-art semantic dependency parser~\citep{peng2017deep}
that considers three types of parts: heads, unlabeled arcs, and labeled arcs.
Let $\sigma(\ledge{i}{j}{\ell})$ denote the index of the arc from $i$ to $j$
with semantic role $\ell$.
In addition to $[0,1]$ constraints,
we constrain that the predictions for labeled arcs sum to the prediction of 
their associated unlabeled arc:
\begin{equation}
\label{eq:sdp_proj}
\begin{split}
\mathcal{P}_{\text{SDP}} 
\left\{\mathbf{p}\in\mathbb{U}^d\left|
\sum_{\ell}p_{\sigma(\ledge{i}{j}{\ell})} =
p_{\sigma(\edge{i}{j})},\forall i\neq j\right.\right\}.
\end{split}
\end{equation}
This ensures that exactly one label is predicted if and only if its arc is present.
The projection onto $\mathcal{P}_{\text{SDP}}$ 
can be solved similarly to Algorithm~\ref{algo:dep_proj}.
We drop the determinism constraint imposed by~\citet{peng2017deep}
in the backward computation.

\section{Experiments}
\label{sec:experiment}
We empirically evaluate our method with two sets of experiments:
using syntactic tree structures in semantic dependency parsing,
and using semantic dependency graphs in sentiment classification.
\subsection{Syntactic-then-Semantic Parsing}
\label{subsec:exp:syntax+sdp}
In this experiment we consider an intermediate syntactic parsing task,
followed by semantic dependency parsing as the end task.
We first briefly review the neural network architectures
for the two models~(\S\ref{subsubsec:exp:syntax+sdp:ach}), 
and then introduce the datasets~(\S\ref{subsubsec:exp:syntax+sdp:data})
and baselines (\S\ref{subsubsec:exp:syntax+sdp:baseline}).

\subsubsection{Architectures}
\label{subsubsec:exp:syntax+sdp:ach}

\paragraph{Syntactic dependency parser.}
For intermediate syntactic dependencies,
we use the unlabeled arc-factored parser of~\citet{kiperwasser2016simple}.
It uses bidirectional LSTMs (BiLSTM) to encode the input,
followed by a multilayer-perceptron (MLP) to score each potential dependency.
One notable modification is that
we replace their use of Chu-Liu/Edmonds' algorithm~\citep{chu-liu1965shortest,edmonds1967optimum} 
with the Eisner algorithm~\citep{eisner1996three,eisner2000bilexical},
since our dataset is in English and mostly projective.

\paragraph{Semantic dependency parser.}
We use the basic model of~\citet{peng2017deep} (denoted as \textsc{NeurboParser}) 
as the end model.
It is a first-order parser, and uses local factors for heads, unlabeled arcs, and labeled arcs.
\textsc{NeurboParser}~does not use syntax.
It first encodes an input sentence with a two-layer BiLSTM,
and then computes part scores with two-layer $\tanh$-MLPs.
Inference is conducted with AD$^3$~\citep{martins2015ad3}.
To add syntactic features to \textsc{NeurboParser}, 
we concatenate a token's contextualized representation
to that of its syntactic head, predicted by the intermediate parser.
Formally, given length-$n$ input sentence, we first run a BiLSTM.
We use the concatenation of the two hidden representations $\bilstm_j=[\fw{j};\bw{j}]$
at each position $j$ as the contextualized token representations.
We then concatenate $\bilstm_j$ with the representation of its head $\bilstm_{\operatorname{HEAD}(j)}$ by%
\begin{align}
\label{eq:concat}
\widetilde{\bilstm}_j=[\bilstm_j;\bilstm_{\operatorname{HEAD}(j)}]
=\left[\bilstm_j;\sum_{i\neq j}\hat{z}_{\sigma(\edge{i}{j})}\ \bilstm_i\right],
\end{align} 
where $\hat{\mathbf{z}}\in\mathbb{B}^{n(n-1)}$ is a binary encoding of the tree structure predicted by
by the intermediate parser.
We then use $\widetilde{\bilstm}_j$ anywhere $\bilstm_j$ would have been used
in \textsc{NeurboParser}.
In backpropagation, we compute $\nabla_{\hat{\mathbf{z}}}L$ with an automatic differentiation toolkit
~\citep[DyNet;][]{dynet}.

We note that this approach can be  generalized to
convolutional neural networks over graphs~\interalia{mou2015discriminative,duvenaud2015convolutional,kipf2016semi},
recurrent neural networks along paths~\interalia{xu2015classifying,roth2016neural}
or dependency trees~\citep{tai2015improved}.
We choose to use concatenations to control the model's complexity,
and thus to better understand which parts of the model work.

We refer the readers to \citet{kiperwasser2016simple} and \citet{peng2017deep} for further details of the parsing models.

\paragraph{Training procedure.}
Following previous work, we minimize 
structured hinge loss~\citep{tsochantaridis2004svm} for both models.
We jointly train both models from scratch,
by randomly sampling an instance from the union of their training data at each step. 
In order to isolate the effect of backpropagation, we do not share any
parameters between the two models.\footnote{
Parameter sharing has proved successful in many related tasks~\interalia{collober2008unified,soggard2016deep,ammar2016many,swayamdipta2016greedy,swayamdipta2017frame},
and could be easily combined with our approach.
}
Implementation details are summarized in the supplementary materials.

\begin{table}[tb]
	\begin{subtable}[tb]{\columnwidth}
		\centering
		\begin{tabulary}{0.47\textwidth}{@{}l  cc c cc@{}} 
			
			\toprule
			& \multicolumn{2}{c}{\textbf{DM}}
			& \phantom{}
			& \multicolumn{2}{c}{\textbf{PSD}}\\

			\cmidrule{2-3}
			\cmidrule{5-6}
			
			\textbf{Model}
			& U$F$ & L$F$ &
			& U$F$ & L$F$ \\
			
			\midrule
			\textsc{NeurboParser}
			& -- &  89.4 &
			& -- &  77.6\\
			
			\textsc{Freda3}
			& -- &  90.4 &
			& -- &  78.5 \\
			
			\midrule
			
			\pipeline
			& 91.8 & 90.8 &
			& 88.4 & 78.1 \\ 
			
			\sa
			& 91.6 & 90.6 &
			& 87.9 & 78.1 \\
			
			\ste
			& 92.0 & 91.1 &
			& \textbf{88.9} & \textbf{78.9} \\
			
			\midrule[.03em]
			\pfgd
			& \textbf{92.4} & \textbf{91.6} &
			& 88.6 & \textbf{78.9} \\ 
			\bottomrule
			
		\end{tabulary}
		\caption{$F_1$ on in-domain test set.}
		\vspace{.5cm}
		\label{tab:syntax_sdp_id}
	\end{subtable}
	\begin{subtable}[tb]{\columnwidth}
		\centering
		\begin{tabulary}{0.47\textwidth}{@{}l  cc c cc@{}} 
			
			\toprule
			& \multicolumn{2}{c}{\textbf{DM}}
			& \phantom{}
			& \multicolumn{2}{c}{\textbf{PSD}}\\

			\cmidrule{2-3}
			\cmidrule{5-6}
			
			\textbf{Model}
			& U$F$ & L$F$ &
			& U$F$ & L$F$ \\
			
			\midrule
			\textsc{NeurboParser}
			& -- &  84.5 &
			& -- &  75.3\\
			
			\textsc{Freda3}
			& -- &  85.3 &
			& -- &  76.4 \\
			
			\midrule
			
			\pipeline 
			& 87.4 & 85.8 &
			& 85.5 & 75.6 \\ 
			
			\sa
			& 87.3 & 85.6 &
			& 84.9 & 75.9 \\
			
			\ste
			& 87.7 & 86.4 &
			& \textbf{85.8} & 76.6\\ 
			
			\midrule[.03em]
			\pfgd
			& \textbf{87.9} & \textbf{86.7} &
			& 85.5 & \textbf{77.1} \\ 
			\bottomrule
			
		\end{tabulary}
		\caption{$F_1$ on out-of-domain test set.}
		\vspace{.1cm}
		\label{tab:syntax_sdp_ood}
	\end{subtable}
	\caption{Semantic dependency parsing performance in both
		unlabeled (U$F$) and labeled (L$F$) $F_1$ scores.
		Bold font indicates the best performance.
		\citet{peng2017deep} does not report U$F$.}
	\label{tab:syntax_sdp}
\end{table}

\subsubsection{Datasets}
\label{subsubsec:exp:syntax+sdp:data}
\begin{compactitem}
	\item
	For semantic dependencies, we use the English dataset from SemEval 2015 Task 18~\citep{oepen2015sdp}. 
	Among the three formalisms provided by the shared task,
	we consider DELPH-IN MRS-derived dependencies (DM) and Prague Semantic Dependencies (PSD).\footnote{We drop the third (PAS)
          because its structure is highly predictable from
          parts-of-speech, making it less interesting.}
	It includes \S00--19 of the WSJ corpus as training data, 
	\S20 and \S21 for development and in-domain test data, 
	resulting in a 33,961/1,692/1,410 train/dev./test split,
	and 1,849 out-of-domain test instances from the Brown corpus.%
	\footnote{The organizers remove, e.g., instances with cyclic graphs, 
		and thus only a subset of the WSJ corpus is included.
		See \citet{oepen2015sdp} for details.}
	\item For syntactic dependencies, we use the Stanford Dependency~\citep{stanforddependency} conversion
	      of the the Penn Treebank WSJ portion~\citep{ptb}. 
	To avoid data leak, we depart from standard split
	and use \S20 and \S21 as development and test data, and the remaining sections as training data.
	The number of training/dev./test instances is 40,265/2,012/1,671.
\end{compactitem}

\subsubsection{Baselines}	
\label{subsubsec:exp:syntax+sdp:baseline}
We compare to the following baselines:
\begin{compactitem}
	\item A pipelined system (\pipeline).
	The pretrained parser achieves 92.9 test unlabeled attachment score (UAS).%
	\footnote{
	Note that this number is not comparable to the parsing literature due to the different split.
	As a sanity check, we found in preliminary experiments that the same parser architecture 
	achieves 93.5 UAS when trained and evaluated with the standard split, close to the results reported by \citet{kiperwasser2016simple}.}
	\item Structured attention networks~\citep[\sa;][]{kim2017structured}. 
	 We use the inside-outside algorithm~\citep{baker1979trainable} to populate $\mathbf{z}$ 
	 with arcs' marginal probabilities, use log-loss as the objective in training the intermediate parser.
	\item The straight-through estimator~\citep[\ste;][]{hinton2012neural}, introduced in \S\ref{subsec:pfgd}.
\end{compactitem}

\subsubsection{Empirical Results}
\label{subsubsec:exp:syntax+sdp:results}
Table~\ref{tab:syntax_sdp} compares the semantic dependency parsing performance of 
\pfgd~to all five baselines.
\textsc{Freda3}~\citep{peng2017deep} is a state-of-the-art variant of \textsc{NeurboParser} 
that is trained using multitask learning to jointly predict three
different semantic dependency graph formalisms.
Like the basic \textsc{NeurboParser} model that we build from, \textsc{Freda3} does not use any syntax.
Strong DM performance is achieved in a more recent work by using
joint learning and an ensemble~\citep{peng2018learning},
which is beyond fair comparisons to the models discussed here.

We found that using syntactic information improves semantic parsing performance:
using pipelined syntactic head features brings 0.5--1.4\% absolute labeled $F_1$ improvement
to \textsc{NeurboParser}.
Such improvements are smaller compared to previous works, where
dependency path and syntactic relation features are included~\citep{almeida2015sdp,ribeyre2015because,zhang2016transition},
indicating the potential to get better performance by using more syntactic information,
which we leave to future work.

Both \ste and \pfgd use hard syntactic features.
By allowing backpropation into the intermediate syntactic parser,
they both consistently outperform \pipeline.
On the other hand, when marginal syntactic tree structures are used,
\sa outperforms \pipeline only on the out-of-domain PSD test set,
and improvements under other cases are not observed.

Compared to \ste, 
\pfgd outperforms \ste on DM by more than 0.3\% absolute labeled $F_1$,
both in-domain and out-of-domain.
For PSD, \pfgd achieves similar performance to \ste on in-domain test set,
but has a 0.5\% absolute labeled $F_1$ improvement on out-of-domain data, 
where syntactic parsing is less accurate.

\subsection{Semantic Dependencies for Sentiment Classification}
\label{subsec:exp:sdp+sst}
Our second experiment uses semantic dependency graphs
to improve sentiment classification performance.
We are not aware of any efficient algorithm that solves
marginal inference for semantic dependency graphs
under determinism constraints, 
so we do not include a comparison to \sa.

\subsubsection{Architectures}
\label{subsubsec:exp:sdp+sst:arch}
Here we use \textsc{NeurboParser} as the intermediate model, as described in
\S\ref{subsubsec:exp:syntax+sdp:ach}, but with no syntactic enhancements.

\paragraph{Sentiment classifier.}
We first introduce a baseline that does not use any structural information.
It learns a one-layer BiLSTM to encode the input sentence,
and then feeds the sum of all hidden states into a two-layer $\operatorname{ReLU}$-MLP.

To use semantic dependency features,
we concatenate a word's BiLSTM-encoded representation
to the averaged representation of its heads, together with the corresponding semantic roles,
similarly to that in Equation~\ref{eq:concat}.%
\footnote{In a well-formed semantic dependency graph,
	a token may have multiple heads. Therefore we use average instead of 
	the sum in Equation~\ref{eq:concat}.}
Then the concatenation is fed into an affine transformation 
followed by a $\operatorname{ReLU}$ activation.
The rest of the model is kept the same as the BiLSTM baseline.

\paragraph{Training procedure.}
We use structured hinge loss to train the semantic dependency parser,
and log-loss for the sentiment classifier.
Due to the discrepancy in the training data size of the two tasks (33K vs.~7K),
we pre-train a semantic dependency parser, and then adopt joint training together with the classifier.
In the joint training stage, we randomly sample 20\% of the semantic dependency training instances each epoch.
Implementations are detailed in the supplementary materials.

\subsubsection{Datasets}
\label{subsubsec:exp:sdp+sst:data}
For semantic dependencies, we use the DM dataset introduced in \S\ref{subsubsec:exp:syntax+sdp:data}.

We consider a binary classification task using the Stanford Sentiment Treebank~\citep{socher2013recursive}.
It consists of roughly 10K movie review sentences from Rotten Tomatoes. 
The full dataset includes a rating on a scale from 1 to 5 for each constituent (including the full sentences), 
resulting in more than 200K instances.
Following previous work~\citep{iyyer2015deep},
we only use full-sentence instances, with neutral instances excluded (3s) and the remaining four rating levels
converted to binary ``positive'' or ``negative'' labels.
This results in a 6,920/872/1,821 train/dev./test split.

\begin{table}[tb]
	\centering
	\begin{tabulary}{\columnwidth}{@{}l c@{}}
		\toprule
		
		\textbf{Model}
		& \textbf{Accuracy (\%)}\\
		
		\midrule
		
		\textsc{BiLSTM}
		& 84.8 \\

		\midrule[.03em]
		
		\pipeline
		& 85.7 \\
		
		\ste 
		& 85.4 \\
		
		\pfgd
		& \textbf{86.3} \\
		
		\bottomrule
	\end{tabulary}
	\caption{Test accuracy of sentiment classification on Stanford Sentiment Treebank.
		Bold font indicates the best performance.}
	\label{tab:sst}
\end{table}

\subsubsection{Empirical Results}
\label{subsubsec:exp:sdp+sst:results}
Table~\ref{tab:sst} compares our \pfgd~method to three baselines.
Pipelined semantic dependency predictions brings 0.9\% 
absolute improvement in classification accuracy,
and \pfgd outperforms all baselines.
In this task \ste~achieves slightly worse performance than a fixed pre-trained \pipeline.

\section{Analysis}
\label{sec:analysis}

We examine here how the intermediate model is affected 
by the end-task training signal.
Is the end-task signal able to ``overrule'' intermediate predictions?

We use the syntactic-then-semantic parsing model~(\S\ref{subsec:exp:syntax+sdp}) as a case study.
Table~\ref{tab:analysis:breakdown} compares a pipelined system 
to one jointly trained using \pfgd.
We consider the development set instances where both syntactic and semantic annotations are available,
and partition them based on whether the two systems'
syntactic predictions agree (\textsc{Same}), or not (\textsc{Diff}).
The second group includes sentences with much lower syntactic parsing
accuracy (91.3 vs.~97.4 UAS), and
\spigot further reduces this to 89.6.
Even though these changes hurt syntactic parsing accuracy, they lead to a 1.1\%
absolute gain in labeled $F_1$ for semantic parsing.
Furthermore, \pfgd has an overall less detrimental effect on the intermediate parser than \ste:
using \pfgd, intermediate dev. parsing UAS drops to 92.5 
from the 92.9 pipelined performance, while \ste reduces it to 91.8.

\begin{table}[tb]
	\centering
	\begin{tabulary}{.96\columnwidth}{@{}c r |c rr@{}} 
		
		\textbf{Split}	& \textbf{\# Sent.}
		&\textbf{Model} &  \textbf{UAS} & \textbf{DM} \\
		
		\hline
		
		
		\multirow{2}{*}{\textsc{Same}} & \multirow{2}{*}{1011}
		&\pipeline & 97.4 & 94.0 \\
		
		& 
		& \pfgd & 97.4 &  94.3 \\
		
		\hline\hline
		\multirow{2}{*}{\textsc{Diff}} & \multirow{2}{*}{681}
		&\pipeline & 91.3 & 88.1 \\
		
		& 
		& \pfgd & 89.6 &  89.2 \\
	\end{tabulary}
	\caption{Syntactic parsing performance (in unlabeled attachment score, UAS) 
		and DM semantic parsing performance (in labeled $F_1$)
		on different groups of the development data. 
		Both systems predict the same syntactic parses
		for instances from \textsc{Same}, and they disagree on instances from \textsc{Diff}~(\S\ref{sec:analysis}). }
	\label{tab:analysis:breakdown}
\end{table}

We then take a detailed look and categorize the changes in intermediate trees 
by their correlations with the semantic graphs.
Specifically, when a modifier $m$'s head is changed from $h$ to $h^\prime$ in the tree,
we consider three cases:
(a) $h^\prime$ is a head of $m$ in the semantic graph;
(b) $h^\prime$ is a modifier of $m$ in the semantic graph;
(c) $h$ is the modifier of $m$ in the semantic graph.
The first two reflect modifications to the syntactic parse that rearrange
semantically linked words to be neighbors.
Under (c), the semantic parser removes a syntactic dependency that reverses
the direction of a semantic dependency.  
These cases account for 17.6\%, 10.9\%, and 12.8\%, respectively
(41.2\% combined) of the total changes.
Making these changes, of course, is complicated, since they often
require \emph{other} modifications to maintain well-formedness of the tree.
Figure~\ref{fig:form} gives an example.
\section{Related Work}
\label{sec:related}

\paragraph{Joint learning in NLP pipelines.}
To avoid cascading errors,
much effort has been devoted to joint decoding
in NLP pipelines~\interalia{habash2005arabi,cohen-07,goldberg2008single,lewis2015joint,zhang2015randomized}.
However, joint inference can sometimes be prohibitively expensive.
Recent advances in representation learning facilitate
exploration in the joint learning of multiple tasks by 
sharing parameters~\interalia{collober2008unified,blitzer-06,finkel-10,zhang2016stack,hashimoto2016joint}.

\paragraph{Differentiable optimization.}
\citet{gould2016differentiating} review the generic approaches to differentiation in bi-level optimization~\citep{bard2006practical,kunishch2013bilevel}.
\citet{amos2017optnet}~extend their efforts to a class of subdifferentiable quadratic programs.
However, they both require that the intermediate objective has an invertible Hessian,
limiting their application in NLP.
In another line of work,
the steps of a gradient-based 
optimization procedure are unrolled into a single computation graph
\citep{stoyanov2011empirical,domke2012generic,goodfellow2013,brakel2013training}.
This comes at a high computational cost 
due to the second-order derivative computation during backpropagation.
Moreover, constrained optimization problems (like many NLP problems)
often require projection steps within the procedure, 
which can be difficult to differentiate through~\citep{belanger2016spen,belanger2017end2end}.

\section{Conclusion}
\label{sec:conclusion}
We presented \pfgd, a novel approach to backpropagating through neural network
architectures that include discrete structured decisions in intermediate layers.
\pfgd devises a proxy for the gradients with respect to
$\argmaxname$'s inputs,
employing a projection that aims to respect the constraints in the intermediate task.
We empirically evaluate our method with two architectures:
a semantic parser with an intermediate syntactic parser,
and a sentiment classifier with an intermediate semantic parser.
Experiments show that \pfgd achieves stronger performance than 
baselines under both settings,
and outperforms state-of-the-art systems on semantic dependency parsing. 
Our implementation is available at \codeurl.

\section*{Acknowledgments}
We thank the ARK, Julian Michael, Minjoon Seo, Eunsol Choi, and Maxwell Forbes for their helpful comments on an earlier version of this work, and the anonymous reviewers for their valuable feedback.
This work was supported in part by
NSF grant IIS-1562364.

\bibliography{acl2018}
\bibliographystyle{acl_natbib}

\pagebreak
\end{document}


\maketitle

\section{Implementation Details}
Our implementation is based on the DyNet toolkit.\footnote{\url{https://github.com/clab/dynet}}
We use part-of-speech tags and lemmas predicted by NLTK.\footnote{\url{http://www.nltk.org/}}

\subsection{Syntactic-then-Semantic Parsing Experiment}
\label{subsec:syntax_sdp}
Each input token is represented as the concatenation a word embedding vector, 
a learned lemma vector, 
and a learned vector for part-of-speech, all updated during training.
In joint training, we apply early-stopping based on 
semantic dependency parsing development performance (in labeled $F_1$).
We do not use mini-batch.
We set the step size $\eta$ for \pfgd to $1$.

\paragraph{Semantic dependency parser.} 
We use the pruning techniques in~\citet{martins2014sdp}, 
and replace their feature-rich model with neural networks~\citep{peng2018learning}.
We observe that the number of parts surviving pruning 
is linear in the sentence length ($5.5\times$ on average),
with  $\sim$99\% recall.

We do not deviate far from the hyperparameter setting in~\citet{peng2017deep},
with the only exception being that we use 50-dimensional lemma and part-of-speech embeddings, instead of 25.

\paragraph{Syntactic dependency parser.}
For the max-margin syntactic parsers used in \pipeline, \ste, and \pfgd,
we use the hyperparameters reported in~\citet{kiperwasser2016simple},
but replace their 125-dimensional BiLSTMs with 200-dimensional ones,
and use 50-dimensional POS embeddings, instead of 25.
We anneal the learning rate at a rate of $0.5$ every 5 epochs. 

For the marginal syntactic parser in \sa,
we follow the use of Adam algorithm~\citep{kingma2014adam}, 
but set a smaller initial learning rate of $5\times 10^{-4}$,
annealed at a rate of $0.5$ every 4 epochs.
The rest of the hyperparameters stay the same as the max-margin parser.

\subsection{Semantic Parsing and Sentiment Classification Experiment}
The model is trained for up to 30 epochs in the joint training stage. 
We apply early-stopping based on 
sentiment classification development accuracy.
For semantic dependency parser, we follow the
hyperparameters described in~\S\ref{subsec:syntax_sdp}.

\paragraph{Sentiment classifier.}
We use 300-dimensional \texttt{GloVe} \citep{pennington2014glove} 
to initialize word embeddings,
fixed during training.
We use a single-layer BiLSTM, followed by a two-layer $\operatorname{ReLU}$-MLP.
Dropout in word embeddings and MLPs is applied, but not in LSTMs.
We use Adam algorithm~\citep{kingma2014adam},
and follow the default procedures by DyNet for optimizer settings and parameter initializations.
An $\ell_2$-penalty of $10^{-6}$ is applied to all weights.
Learning rate is annealed at a rate of $0.5$ every 5 epochs.
We use mini-batches of 32,
and clip the $\ell_2$-norm of gradients to 5~\citep{graves2013generating}.
We set the step size $\eta$ for \pfgd to $\frac{5}{32}$.
We explore the same set of hyperparameters based on development performance for all compared models, 
summarized in Table~\ref{tb:hyperparameters}.

\begin{table}
	\centering
	\begin{tabulary}{\columnwidth}{@{}l r@{}}
		
		\toprule
		
		\textbf{Hyperparameter} & \textbf{Values}\\
		\midrule
		MLP dimension & $\{100, 150, 200, 250, 300\}$ \\
		BiLSTM dimension & $\{100, 150, 200, 250, 300\}$\\
		Embedding dropout &  $\{0.2, 0.3, 0.4, 0.5\}$\\
		MLP dropout & $\{0.0, 0.1, 0.2, 0.3, 0.4\}$\\ 
		
		\bottomrule
	\end{tabulary}
	\caption{Hyperparameters explored in sentiment classification experiments.}
	\label{tb:hyperparameters}
\end{table}

\bibliography{acl2018}
\bibliographystyle{acl_natbib}